\documentclass[10pt,twocolumn,letterpaper]{article}

\usepackage[pagenumbers]{cvpr} % To force page numbers, e.g. for an arXiv version

\definecolor{cvprblue}{rgb}{0.21,0.49,0.74}
\usepackage[pagebackref,breaklinks,colorlinks,allcolors=cvprblue]{hyperref}
\usepackage{multirow}
\usepackage{array}
\usepackage{CJKutf8}
\def\paperID{10674} % *** Enter the Paper ID here
\def\confName{CVPR}
\def\confYear{2025}

\title{Hybrid Discriminative Attribute-Object Embedding Network for Compositional Zero-Shot Learning}

\author{
Yang~Liu$^{1}$, Xinshuo~Wang$^{1}$, Jiale~Du$^{1}$, Xinbo~Gao$^{1,2}$, Jungong~Han$^{3}$ \\
$^{1}$Xidian University, Xi'an, China \\
$^{2}$Chongqing University of Posts and Telecommunications, Chongqing, China \\
$^{3}$Tsinghua University, Beijing, China \\
{\tt\small yangl@xidian.edu.cn, wangxinshuo2003@163.com, 23011211070@stu.xidian.edu.cn, }\\{\tt\small xbgao@mail.xidian.edu.cn, jungonghan77@gmail.com}
}

\begin{document}
\begin{CJK}{UTF8}{gbsn}
\maketitle
\begin{abstract}
Compositional Zero-Shot Learning (CZSL) recognizes new combinations by learning from known attribute-object pairs. However, the main challenge of this task lies in the complex interactions between attributes and object visual representations, which lead to significant differences in images. In addition, the long-tail label distribution in the real world makes the recognition task more complicated. To address these problems, we propose a novel method, named Hybrid Discriminative Attribute-Object Embedding (HDA-OE) network. To increase the variability of training data, HDA-OE introduces an attribute-driven data synthesis (ADDS) module. ADDS generates new samples with diverse attribute labels by combining multiple attributes of the same object. By expanding the attribute space in the dataset, the model is encouraged to learn and distinguish subtle differences between attributes. To further improve the discriminative ability of the model, HDA-OE introduces the subclass-driven discriminative embedding (SDDE) module, which enhances the subclass discriminative ability of the encoding by embedding subclass information in a fine-grained manner, helping to capture the complex dependencies between attributes and object visual features. The proposed model has been evaluated on three benchmark datasets, and the results verify its effectiveness and reliability.
\end{abstract}
\section{Introduction}
Humans can easily recognize new combinations of objects and attributes, like imagining a blue horse, by reasoning about different object aspects and generalizing knowledge to unseen combinations. In Compositional Zero-Shot Learning (CZSL) \cite{purushwalkam2019task,mancini2022learning,li2023distilled,wang2023bi}, the goal is to predict unseen combinations of objects and attributes after learning from known classes and their descriptions. For instance, after learning about ``Red Goldfish" and ``Yellow Koi", the model can recognize a ``Red Koi". This task is challenging due to variations in shapes, colors, and textures across different attribute-object combinations.
\begin{figure}[!t]
    \centering
    \includegraphics[width = 3.2in]{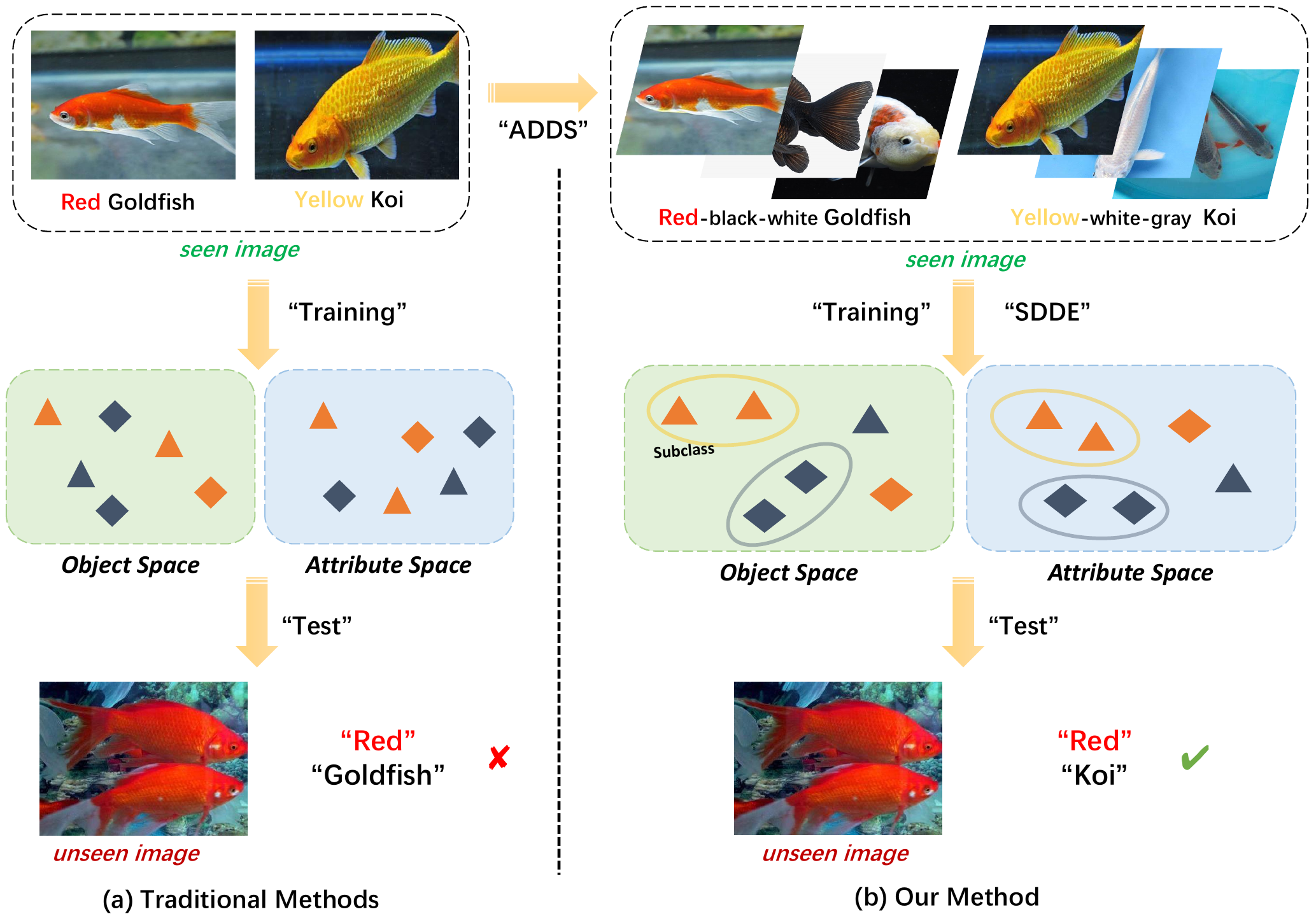}
    \caption{(a) Traditional methods: Recognize unseen images by learning from known combinations. (b) Our method: Image samples are expanded across multiple layers, after which our visual features are deconstructed and mapped into the corresponding spaces, ultimately converging to form category prototypes. These prototypes serve as a basis for reassembling and predicting new combinations.}
    \label{figure1}
\end{figure}
Traditional methods \cite{misra2017red,naeem2021learning,nagarajan2018attributes,mancini2022learning} treat each attribute-object combination independently and classify each pair as a distinct category, disregarding the relationships between combinations. For example, Misra \etal \cite{misra2017red} focus on distinguishing unrelated pairs but struggle with variability within a class, while Nagarajan \etal \cite{nagarajan2018attributes} separate attribute and object features, overlooking their interactions. These methods fail to capture subtle subclass distinctions and perform poorly on CZSL datasets. 

%To address these issues, we introduce a novel embedding module and a composite embedding generated through consistency constraints. This method produces virtual embeddings with enhanced subclass discrimination, allowing the model to better differentiate between fine variations in attribute-object combinations. By incorporating contrastive learning, our method maintains distinctions between combinations in the disentangled embeddings.

To address these issues, we introduce the Subclass Driven Discriminative Embedding (SDDE) method. This method enhances the encoding's sensitivity to sub-class distinctions by performing fine-grained sub-class embeddings within each category. In this way, subclasses in the embedding space are clustered, which facilitates the model to accurately classify different combinations during recognition. As shown in Figure \ref{figure1}, during the training process, the features are grouped into subclasses after being processed by the SDDE, with similar features being clustered together. Ultimately, SDDE enables the model to better capture subtle changes in attribute-object combinations and improve its discrimination ability.

In addition to visual differences between subclasses, real-world image hybridity and the long-tail distribution of labels present significant challenges in CZSL. Some methods such as Saini \etal \cite{kim2023hierarchical} model each combination separately, limiting generalization. Wang \etal \cite{wang2023learning} rely on supervised learning, which lacks the ability to generalize to unseen combinations, while Kim \etal \cite{Saini_2022_CVPR} attempt to encode attributes and objects independently but struggle with complex interactions and subclass distinctions.

%To address these issues, we propose a dataset combination strategy to enhance diversity. By creating a new dataset with variations of the same object and combining it with existing data, our approach captures complex interactions between attribute-object pairs, improving generalization and recognition accuracy. Embedding experts use a dynamic weighting mechanism to adjust attributes and objects, making the model more flexible and better suited for capturing real-world attribute-object distributions. This strategy enriches the training data and improves feature representation, leading to stronger performance in recognizing new combinations.

To address these issues, we propose the Attribute-Driven Data Synthesis (ADDS) method to enhance data diversity. ADDS expands the attribute space in the training set by combining different attributes  with the same object to generate new samples with significant visual differences. As shown in Figure \ref{figure1}, the ``Red Goldfish" is extended to  ``Red-black-white Goldfish". This not only increases the diversity of the data, but also helps the model to perform more effective reasoning and classification when faced with new combinations or unknown attribute-object pairs, especially showing stronger robustness when dealing with highly varied and rare combinations.

Our main contributions can be summarized:
\begin{itemize}
  \item We propose a Hybrid Discriminative Attribute-Object Embedding (HDA-OE) network that balances data distribution and improves generalization by broadening the attribute space in the training set.
  \item We propose a subclass-driven discriminative embedding module to strengthen subclass differentiation between attribute-object combinations, significantly enhancing the model's discriminative power.
  \item We conduct extensive experiments on three challenging benchmark datasets (\ie, MIT States, UT Zappos, and C-GQA) under both open-world and closed-world CZSL settings. Results show that our HDA-OE achieves significant improvements and new state-of-the-art results.
\end{itemize}

\section{Related Work}
\subsection{Zero-Shot Learning}
Zero-shot learning (ZSL) classifies objects in unseen categories by transferring knowledge from seen categories, using semantic information like attributes, text descriptions, or word embeddings, without relying solely on visual data \cite{li2021attribute, yang2021adaptive}. This approach is highly adaptable in recognizing new categories \cite{xie2019attentive}. The first is embedding-based methods, such as Zhang \etal \cite{zhang2020towards}, which emphasize embedding spaces that ensure intra-class cohesion and inter-class separation. Other studies, such as Bi-VAEGAN \cite{wang2023bi}, explored alternative embedding spaces to effectively connect visible and invisible elements. Techniques such as second-order pooling \cite{lampert2013attribute} and prototype learning \cite{zhang2017learning} further refine image representations, while generative methods such as conditional VAEs decompose images into semantically meaningful components \cite{li2021disentangled,li2021generalized}. In addition, graph convolutional networks (GCNs) \cite{kipf2016semi,kampffmeyer2019rethinking, wang2018zero} show good promise by using knowledge graphs to predict unseen categories and making improvements to alleviate issues such as Laplacian smoothing \cite{kampffmeyer2019rethinking, li2018deeper}. Together, these approaches emphasize the importance of semantic feature integration, making ZSL a powerful tool for combinational and zero-shot learning tasks.
\subsection{Compositional Zero-Shot Learning}
%The problem statement of compositional zero-shot learning (CZSL) \cite{nagarajan2018attributes, Atzmon2020causal, li2020symmetry} shares some similarities with traditional zero-shot learning (ZSL). However, while ZSL primarily aims to recognize unseen objects, CZSL focuses on recognizing unseen combinations of states and objects. CZSL builds upon the foundation of ZSL by delving deeper into the various aspects of sample combinations.
Compositional Zero-Shot Learning (CZSL) \cite{misra2017red,nagarajan2018attributes, Atzmon2020causal, li2020symmetry, jiang2024mutual}
focuses on identifying unseen combinations of states and objects by examining various aspects of sample combinations.
Existing methods fall into two categories: the first maps inputs to combination space for classification, using two classifiers to independently recognize object and state class prototypes. For instance, Chen \etal \cite{chen2014inferring} proposed a tensor decomposition method to infer unseen object-state pairs using sparse class-specific SVM classifiers trained on visible components. Nagarajan \etal \cite{nagarajan2018attributes} suggested that the transformation of object features within the combination is a linear function of state features. Atzmon \etal \cite{Atzmon2020causal} proposed a discriminative model to ensure conditional independence between state and object recognition. However, due to significant visual deviations between objects and states, these methods often struggle in practical applications. The second category focuses on learning a joint representation of state-object combinations for classification. Recently, Naeem \etal \cite{naeem2021learning} introduced a GCN-based model to capture dependencies between objects and states, addressing CZSL challenges. SymNet \cite{li2020symmetry} utilized the symmetric relationship between states and objects to filter out impossible combinations and improve prototype quality. A contrastive learning method \cite{li2022siamese} enhanced generalization for new combinations by isolating class prototypes, while Khan \etal \cite{hao2023learning} employed self-attention to capture component interdependencies, refining label embeddings for better differentiation.

\subsection{Overcoming Training Data Limitations}
To further improve model performance, many methods \cite{lin2017focal, zhou2017fine, jiang2024mutual} focus on training data and address issues such as sample imbalance and data mixing between classes. For example, Redmon \etal \cite{redmon2017yolo9000} divided labels into levels based on their structural links, then augments the data at each level to preserve balance across categories. Zhou \etal \cite{zhou2017fine} proposed an active incremental learning method to encourage the model to prioritize learning more difficult classes, thus mitigating the impact of simpler sample domains. Lin \etal \cite{lin2017focal} combined the difficulty of each sample with the objective function to adaptively assess sample difficulty during each iteration. Jiang \etal \cite{jiang2024mutual} evaluated the visual bias of two components to assess their imbalance and reweights the training process of CZSL using this imbalance information.
Different from traditional methods, we construct a new dataset that is related to the original one but has certain differences. These two datasets are then combined to form the database required for our model training. This strategy not only introduces greater diversity into the training data but also exposes the model to a wider variety of images and attribute combinations during training.

\section{Approach}
\begin{figure*}[!t]
\centering
    \includegraphics[width = \textwidth]{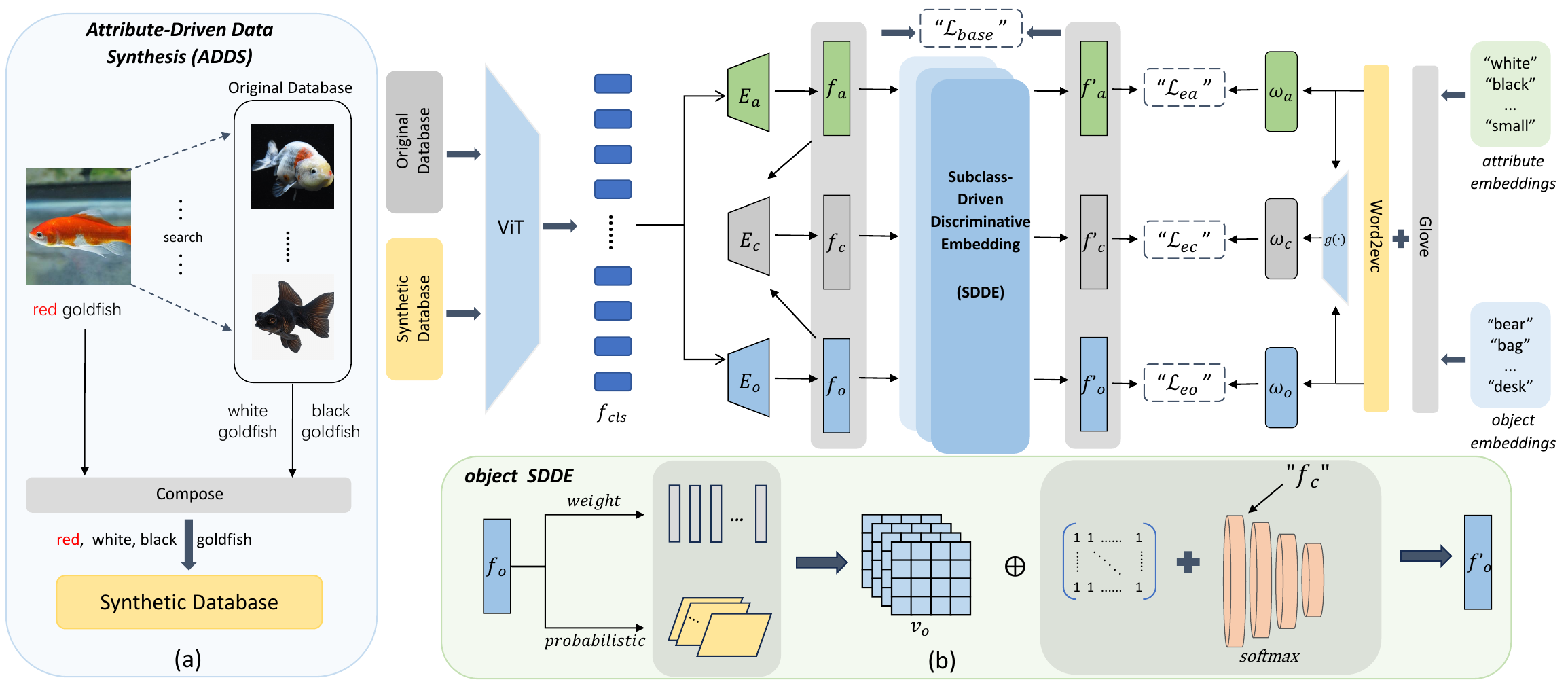}
    \caption{An overview of the proposed approach. We generate the target database by Attribute-Driven Data Synthesis (ADDS) (as shown in (a)). Then, we decompose the encoded visual features (\emph{i.e.}, $f_{cls}$) into their corresponding attribute and object feature embeddings using a traditional disentanglement architecture. A series of target feature embeddings with enhanced discriminative power will be synthesized through Subclass-Driven Discriminative Embedding (SDDE) (as shown in (b)). Both the target feature embeddings and the original feature embeddings are projected into a shared space to achieve semantic alignment.}
\label{network}
\end{figure*}

%In this section, a detailed description of our approach is provided. We show the database construction part and then present the content of feature extraction encoding. Subsequently, we introduce the detailed implementation process of the embedding expert module in detail. The overall architecture is shown in Figure \ref{network}.
The overall architecture is illustrated in Figure \ref{network}. We begin with the database construction, where a new hybrid database and its embeddings are generated by combining multiple datasets. Next, we present the feature extraction encoding, which decomposes encoded visual features using a traditional disentanglement framework. Following this, we elaborate on the implementation of the embedding expert module, which uses contrastive learning to align the generated virtual encoding with the original embedding, producing a virtual embedding with improved subclass discrimination.

\subsection{Problem Definition}
Let the set of possible attributes in the dataset be $\mathcal{A}=\{a_1,a_2,...,a_m\}$, and the set of possible objects be $\mathcal{O}=\{o_1,o_2,...,o_n\}$. By combining attributes and objects, we can form all possible attribute-object pairs, creating a set $Y = \mathcal{A} \times \mathcal{O}$, and the total number of compositions can be calculated as $|Y|=m\cdot n$.
In the CZSL setting, the set $Y$ should be split into two disjoint parts, the visible component set $Y_{s}$ and the invisible component set $Y_{u}$, where $|Y_{s}| + |Y_{u}| \leq |Y|$.
%We partition $\mathcal{Y}$ into two classes: the visible class $y_{s}$ and the invisible class $y_{u}$, where $y_{s} \cap y_{u} = \emptyset$.
During model training, we utilize samples from the visible class $Y_{s}$, denoted as $D_{tr} = \{(X_{s}, Y_{s})\}$. Assume that $X$ is the set of images corresponding to $Y$, and $X_{s}$ corresponds to $Y_{s}$.
For testing, we define two setups based on the range of the output label space: CW-CZSL (Closed-world Compositional Zero-shot Learning) and OW-CZSL (Open-world Compositional Zero-shot Learning). In CW-CZSL, the test set $D_{t} = \{(X_{t}, Y_{t})\}$ comprises samples from the visible class $Y_{s}$ and all samples from the invisible class $Y_{u}$ , $Y_{t} = Y_{u} \cup Y_{st}$, where $Y_{st}$ belongs to $Y_{s}$. In OW-CZSL, the output space extends to all potential attribute-object pairs, \emph{i.e.}, $Y_{t}=Y$. This setup enables evaluating the model's ability to generalize to unseen categories, thereby enhancing its performance in real-world applications.

\subsection{Baseline Framework}
When presented with an input image $x$, we leverage the ViT backbone network to extract its visual features, denoted as $f_{cls}$, representing the visual content of $x$.
%Initially, the image undergoes segmentation into patches, which are then fed into the VIT to yield a $[cls]$ token representation. Subsequently,
The resulting feature blocks are forwarded to two encoders, namely  $E_{a}$ (attribute encoder) and $E_{o}$ (object encoder), to derive its visual embedding. Each encoder is tasked with encoding $x$ to generate an embedding in its respective domain, resulting in $f_{a}$ and $f_{o}$:
\begin{equation}
f_{o} = Norm(E_{o}(f_{cls}))\  ,\  f_{a} = Norm(E_{a}(f_{cls})),
\end{equation}
where $Norm(\cdot)$ stands for normalization. By employing $E_{c}$ (composite encoder) to merge $f_{a}$ and $f_{o}$, we obtain $f_{c}$, the combined visual embedding:
\begin{equation}
f_{c} = Norm(E_{c}(Concat[f_{o},f_{a}])).
\end{equation}

Additionally, we generate the requisite word embeddings $w_{a}$ and $w_{o}$ utilizing $Glove$ and $Word2vec$ dual word vectors. We approximate the synthetic embedding $w_{c}$ by projecting the concatenated word vectors into the joint space.
\begin{equation}
w_{c} = g(Concat[w_{o},w_{a}]),
\end{equation}
where $g(\cdot)$ is a label embedding network, consisting of 3 FC layers and ReLU activation function.
In the context of the object domain, to amalgamate object semantic information, we generate predictions by computing the cosine similarity between cosine visual embeddings $f$ and word embeddings $w$.
%The formula is shown below:
%\begin{equation}
%C = Cos(f,w) = \frac{f^T \cdot w}{\|f\|_2 \cdot \|w\|_2}.
%\end{equation}
%In this context, $ \| \cdot \|_{2} $ denotes the Euclidean norm of a vector.

We introduce three separate cross-entropy loss functions to maximize the recognition probability in each of these spaces, thereby optimizing the model across all three domains. The loss functions are defined as follows:
\begin{equation}
\mathcal{L}_o = - \sum_{o \in \mathcal{\mathcal{O}}} \log \frac{\exp \left( \frac{1}{\tau} \cdot \mathcal{C} (f_o, w_o ) \right)}{\sum_{o' \in \mathcal{O}} \exp \left( \frac{1}{\tau} \cdot \mathcal{C} (f_o, w_{o'} ) \right)},
\end{equation}
where \(\tau\) is the temperature factor, and \(\mathcal{C}(f, w) = \text{cos}(f, w) = \frac{f^T \cdot w}{\|f\|_2 \cdot \|w\|_2}\), using \(\|\cdot\|_2\) to represent the Euclidean norm of the vector. The loss functions for the attribute space and the attribute-object space are formulated similarly to the object space loss function. The overall training loss is a linear combination of the losses from these three spaces:
\begin{equation}
\mathcal{L}_{base}=\mathcal{L}_{a}+\mathcal{L}_{o}+\mathcal{L}_{c}.
\end{equation}

\subsection{Attribute-Driven Data Synthesis (ADDS)}
In our database construction strategy, we adopted a hybrid approach to enhance recognition accuracy. Initially, we established a widely recognized database, termed $D_{A}$, which comprises image data paired with corresponding attribute-object information. Within $D_{A}$, we randomly select an image $x_{a}$, and leverage its attributes and object details to choose images sharing the same object but exhibiting differing attributes. If a given object possesses only one attribute, we directly select an image from those associated with the object; otherwise, we select a new attribute based on its distribution and subsequently choose an image featuring the selected attribute associated with the object. If the selected image does not align with the given attributes, we iteratively reselect based on weights until a congruent image is found. The weight calculation formula is as follows:
\begin{equation}
weight_i = \frac{1/count_i}{\sum_j (1/count_j)},
\end{equation}
where $count_{i}$ denotes the occurrence count of each attribute in images of a given object.

This approach yields database $D_{B}$, housing image datasets akin to those in $D_{A}$ but bearing different attribute labels. Through connector $E_{d}$, we amalgamate database $D_{A}$ and database $D_{B}$ to form a new database:
\begin{equation}
D_{C}= E_{d}([D_{A}, D_{B}]).
\end{equation}
%Within this novel database, for any image x, its word embedding $w_{C}$ is generated from word embeddings $w_{A}$ and $w_{B}$ from database $D_{A}$  and $D_{B}$, that is:
%\begin{equation}
%w_{C} = MLP(Concat[w_{A}, w_{B}]).
%\end{equation}

%Such a methodology aids in infusing diversity into the dataset and furnishes a broader data distribution, thereby supporting more comprehensive and precise model training and evaluation.
The connector \( E_d \) reorganizes the images and attribute-object combinations from both databases, generating new combinations through a combination of connections and multi-layer perceptrons. This mechanism allows the newly created database \( D_C \) to enhance the model's ability to learn attribute-object relationships, particularly in the context of zero-shot learning tasks. %By expanding the dataset in this manner, the operation significantly boosts the model's generalization capability, enabling it to handle previously unseen attribute-object pairs more effectively. 
By generating new attribute-object combinations based on attributes and leveraging a weighted selection process, ADDS enhances both the diversity and representativeness of training data, which in turn boosts the model’s performance on unseen data.

\subsection{Subclass-Driven Discriminative Embedding (SDDE)}
In this section, to enhance the discrimination between different concept pairs in classification learning, we propose the Subclass-Driven Discriminative Embedding (SDDE) module. Using the object embedding expert as an example, Taking the object SDDE as an example, we combine the input object features $f_o$ through a set of probabilistic operations and weight operations to finally obtain the virtual code $v_o$ of the object domain. Next, we combine the decoding of attribute embedding $f_{a}$ and object embedding $f_{o}$ to generate the attribute-object domain virtual coding $v_{c}$. The attribute SDDE obtains $v_a$ in the same way. This virtual coding contains richer subcategory discrimination information compared to directly obtained embeddings, enabling better preservation of subcategory distinctions. Consequently, this approach facilitates the differentiation of various concept pairs.

After obtaining the virtual encoding of the attribute domain and object domain, we can extract classifier-sensitive object and attribute embeddings for concept pair recognition. We start by using the synthetic embedding $f_c$ as the reference point to adjust the virtual encoding $v_a$ and $v_o$, ensuring that our virtual embedding can be effectively mapped back to the corresponding subclass clustering center. Considering the object virtual encoding $v_o$, we first normalize the synthetic embedding to derive subclass attention. Then, we apply the Hadamard product to jointly process the subclass attention and $v_o$. Finally, we combine this result with $f_c$ to obtain a new object embedding $f_o'$:
\begin{equation}
f_o' = v_o + v_o \otimes softmax(f_c).
\end{equation}

This new object embedding $f_o'$ possesses stronger discrimination and generalization capabilities compared to the original $f_o$. After performing the same operation on the attribute domain, we obtained the new attribute embedding $f_{a}'$. Subsequently, we utilized the synthetic encoder $E_c$ and the label embedding network $g(\cdot)$ mentioned earlier to combine $f_{a}'$ with $f_{o}'$, and then combined the result with the virtual encoding $v_c$ once more to obtain a new combined embedding $f_c'$:
\begin{equation}
f'_{c} = g(Concat[E_{c}(Concat[f'_{o},f'_{a}]),v_c]).
\end{equation}

Taking the object domain as an example, we calculate the cosine similarity between $f_{o}'$ and $w_{o}$ to obtain the prediction. Then, we select the combination that yields the highest prediction score, resulting in a new classification loss:
\begin{equation}
\mathcal{L}_{eo} = - \sum_{o \in \mathcal{\mathcal{O}}} \log \frac{\exp \left( \frac{1}{\tau} \cdot \mathcal{C} (f'_o, w_o ) \right)}{\sum_{o' \in \mathcal{O}} \exp \left( \frac{1}{\tau} \cdot \mathcal{C} (f'_o, w_{o'} ) \right)}.
\end{equation}

We combine losses in object space, attribute space, and attribute-object space using a linear function to obtain the embedding loss:
\begin{equation}
\mathcal{L}_{emd}=\mathcal{L}_{ea}+\mathcal{L}_{eo}+\mathcal{L}_{ec}.
\end{equation}

Finally, the total contrast loss $\mathcal{L}_{total}$ can be expressed as follows:
\begin{equation}
\mathcal{L}_{total}=\alpha{\mathcal{L}_{base}} + \beta{\mathcal{L}_{emd}},
\end{equation}
where $\alpha$ and $\beta$ are weighting coefficients utilized to balance the influence of each loss function respectively.

During validation and testing, we aggregate similarities between cosine visual embeddings $f$ and word embeddings $w$, using this as a feasibility score for images and labels. The overall feasibility score $C(a,o)$ is calculated as follows:
\begin{equation}
\begin{scriptsize}
{C}(y=(a,o)) = \mathcal{C} (f'_a, w_a ) + \mathcal{C} (f'_o, w_o ) + \mathcal{C} (f'_c, w_c ).
\end{scriptsize}
\end{equation}
\section{Experiment}
\subsection{Datasets}
We utilize three standard datasets for the zero-shot composition learning (CZSL): UT-Zappos\cite{yu2014fine}, MIT-States\cite{isola2015discovering}, and C-GQA\cite{naeem2021learning} datasets. The details and data partitioning of these datasets are outlined in Table \ref{data}.
For the Mit-States\cite{isola2015discovering}, the output space comprises 1262 visible components and 300/400 invisible components (for validation/testing) in closed-world. We encompass all possible 28,175 compositions within the search space in open-world.
%UT Zappos \cite{yu2014fine} is a medium-sized footwear dataset containing 16 attribute categories and 12 object categories.
The output space for the UT-Zappos \cite{yu2014fine} is restricted to 83 observed configurations in the closed-world context. Forty-one unseen configurations are added for testing and validation, respectively. Despite 40\% (76 out of 192) of possible combinations not being present in any split of the dataset, we account for them within the open-world environment.
Lastly, C-GQA \cite{naeem2021learning} outputs a space of 5592 training components in a closed-world setting, and generates a search space of 278362 components in an open-world setting.
%Lastly, C-GQA \cite{naeem2021learning} stands as a large-scale benchmark dataset, featuring 674 object categories and 413 state categories. In the closed-world environment, the output space encompasses 5592 training compositions. In the open-world environment, it generates a search space comprising 278362 compositions.
%It's evident that the C-GQA dataset presents greater challenges compared to other benchmark datasets.

\begin{table}[!t]
    \centering
    \scriptsize
    \resizebox{\linewidth}{!}{
    \begin{tabular}{ccc|cc|ccc|ccc}
    \hline\hline
        \multicolumn{3}{c|}{}  &\multicolumn{2}{c|}{\textbf{Training}}&	\multicolumn{3}{c|}{\textbf{Validation}}&	\multicolumn{3}{c}{\textbf{Test}}\\
        \textbf{Dataset}&	a	&o&	sp	&	i&	sp	&up&	i&	sp	&up&	i\\ \hline
        UT-Zappos&   16 & 12 &83 & 23k & 15 &15 &  3k & 18 & 18 & 3k\\
        C-GQA&   413&  674&  5592&  27k&  1252&  1040 & 7k& 888&  923 & 5k \\
        MIT-States&  115 &245 &1262& 30k&300& 300& 10k& 400& 400& 13k\\ \hline
    \end{tabular}}
    \caption{Dataset statistics for CZSL: UT-Zappos, MIT States and C-GQA.}
    \label{data}
\end{table}

\subsection{Evaluation Metrics}
Considering our emphasis on a wide range of scenarios and the model's inherent bias towards predicting unseen components, our evaluation scheme follows the approach outlined in \cite{mancini2021open, purushwalkam2019task}. To balance the accuracy between visible and unseen combinations, we introduce a bias factor that favors unseen combinations, offsetting the inherent advantage of visible combinations. During testing, we adjust this bias towards visible combinations to optimize various metrics, aiming to achieve the best seen accuracy (S), the best unseen accuracy (U), the best harmonic mean (HM), and the area under the curve (AUC).
Specifically, we primarily focus on two overall metrics: AUC and HM.
\subsection{Implementation Details}
Similar to the approach in \cite{kim2023hierarchical}, our image features are extracted from a Vision Transformer (ViT)\cite{dosovitskiy2020image} pre-trained on ImageNet \cite{deng2009imagenet}, and visual embeddings are learned based on these features. It is worth noting that we do not use the CLIP \cite{radford2021learning} for training. For the three benchmark datasets, we use 300-dimensional $GloVe$ \cite{pennington2014glove} to initialize the embedding function. We generate 300-dimensional prototype vectors through a fully connected (FC) layer for both $E_o$ and $E_a$. Following \cite{naeem2021learning}, we use a three-layer multi-layer perceptron (MLP) with layer normalization \cite{ba2016layer} and dropout \cite{srivastava2014dropout} for $E_o$ and $E_a$.
%Our label embedding network, $g(\cdot)$, contains three FC layers and two dropout layers with ReLU activation functions, which also generate 600-dimensional embedding vectors to match the word vector embeddings.
The model is trained end-to-end using the Adam optimizer \cite{kendall2018multi}, with a learning rate of $5e-5$, decaying by a factor of 0.1 every 10 epochs, and a temperature parameter $\tau$ set to 0.05. And we set the value of $\alpha : \beta$ to 2:1.

\begin{table*}[!t]
    \centering
    \resizebox{0.9\textwidth}{!}{
        \begin{tabular}{c|c|cccc|cccc|cccc}
        \hline
        \multirow{2}{*}{Closed-world Models} & \multirow{2}{*}{Backbone}&\multicolumn{4}{c|}{UT-Zappos 50K}&	\multicolumn{4}{c|}{C-GQA}&	\multicolumn{4}{c}{MIT-States}\\
        \cline{3-14}
        & &AUC&HM&S&	U&AUC&HM&S&	U&AUC&HM&S&	U\\
        \hline
        OADis~\cite{Saini_2022_CVPR} &\multirow{4}{*}{Resnet}& 30.0 &44.4 &59.5 &65.5 &2.9 &13.1 &30.5 &12.5 &5.9 &18.9 &31.1 &25.6\\
        CANet~\cite{wang2023learning}&  &33.1&47.3&61.0&66.3&3.3&14.5&30.0&13.2&5.4&17.9&29.0&26.2\\
        PSC-VD~\cite{li2024agree}& &33.1 &48.5 &64.8 &65.9 &3.8 &13.0 &29.2 &13.2 &6.4 &20.4 &30.3 &28.3\\
        CSCNet~\cite{zhang2024cscnet}& &-&-&-&-&3.4&14.4 &30.4 &13.4&5.7 &18.4 &30.0 &26.2\\
        \hline
        IVR~\cite{zhang2022learning} &\multirow{10}{*}{ViT}& 34.1 & 48.9 & 61.4 & 68.3 & 2.2 & 10.9 & 27.1 & 10.1 & 5.3 & 18.3 & 26.8 & 28.1 \\
        CompCos~\cite{mancini2021open} & & 31.8 & 48.1 & 58.8 & 63.8 & 2.9 & 12.8 & 30.8 & 12.3 & 4.5 & 16.5 & 25.4 & 24.6 \\
        GraphEmbed~\cite{naeem2021learning}&  & 34.5 & 48.6 & 61.6 & \textbf{70.0} & 3.9 & 15.0 & 32.4 & 15.0 & 5.2 & 18.2 & 31.5 & 28.8 \\
        SCEN~\cite{li2022siamese} & & 31.0 & 46.8 & 65.8 & 62.9 & 3.5 & 14.5 & 31.8 & 13.4 & 4.7 & 7.7 & 33.1 & 27.4 \\
        Co-CGE~\cite{mancini2022learning}&  & 30.8 & 44.6 & 60.9 & 62.6 & 3.7 & 14.7 & 31.6 & 14.4 & 6.7 & 20.1 & 32.1 & 28.4 \\
        DLM~\cite{hu2024dynamic}& &37.7 &52.1 &\textbf{66.5} &68.1 &3.3 &14.8 &30.7 &14.5 &5.8 &19.2 &30.7 &26.6\\
        ADE~\cite{hao2023learning} & &35.1&51.1&63.0&64.3&5.2&18.0&35.0&17.7&7.4&21.2&34.2&28.4\\
        OADis~\cite{Saini_2022_CVPR}& & 32.7 & 46.9 & 60.7 & 66.7 & 3.8 & 14.8 & 33.1 & 14.3 & 5.6 & 17.7 & 32.3 & 27.9 \\
        COT~\cite{kim2023hierarchical} & &34.8 &48.7 &60.8 &64.9 &5.1 &17.5 &34.0 &18.8 &7.8 &23.2 &34.8 &31.5\\
         \hline
        HDA-OE& ViT&\textbf{38.4}&\textbf{54.0}&63.4&68.7&\textbf{6.8}&\textbf{21.1}&\textbf{38.8}&\textbf{20.5}&\textbf{10.9}&\textbf{26.0}&\textbf{39.5}&\textbf{36.1}\\
        \hline
        \end{tabular}
    }
    \caption{Closed-world results on three datasets. We report the area under curve (AUC), the best harmonic
    mean (HM), the best seen accuracy (Seen), and the best unseen accuracy (Unseen) of the unseen-seen
    accuracy curve under the closed-world setting. HM and AUC are the core CZSL metrics.}
\label{tableclose}
\end{table*}

\begin{table*}[!t]
    \centering
    \resizebox{0.9\textwidth}{!}{
        \begin{tabular}{c|c|cccc|cccc|cccc}
        \hline
        \multirow{2}{*}{Open-world Models} & \multirow{2}{*}{Backbone}&\multicolumn{4}{c|}{UT-Zappos 50K}&	\multicolumn{4}{c|}{C-GQA}&	\multicolumn{4}{c}{MIT-States}\\
        \cline{3-14}
        & &AUC&HM&S&	U&AUC&HM&S&	U&AUC&HM&S&	U\\
        \hline
        CANet~\cite{wang2023learning} &\multirow{3}{*}{Resnet}&22.1&38.7&58.7&46.0&0.4&3.2&27.3&1.9&1.2&6.6&25.3&6.7\\
        SAD-SP~\cite{liu2023simple}& &28.4&44.0&63.1&54.7&1.0&5.9&31.0&3.9&1.4&7.8&29.1&7.6\\
        ProCC~\cite{huo2024procc}& &22.4 &39.9 &62.2 &48.0 &0.5 &3.8 &29.0 &2.6 &1.9 &7.8 &27.6 &10.6\\
        \hline
        IVR~\cite{zhang2022learning}&\multirow{10}{*}{ViT} & 24.9 & 41.9 & 59.6 & 50.2 & 0.9 & 5.7 & 30.6 & 3.9 & 4.4 & 17.2 & 25.4 & 23.6 \\
        CompCos~\cite{mancini2021open} & & 20.7 & 36.0 & 58.2 & 46.0 & 0.7 & 4.4 & 32.8 & 2.8 & 4.0 & 16.7 & 24.9 & 21.7 \\
        GraphEmbed~\cite{naeem2021learning}& & 23.5 & 40.1 & 60.6 & 47.1 & 0.8 & 4.9 & 32.8 & 3.2 & 4.3 & 16.8 & 26.3 & 25.0 \\
        SCEN~\cite{li2022siamese}& & 22.5 & 38.1 & \textbf{64.8} & 47.5 & 0.3 & 2.5 & 29.5 & 1.5 & 4.1 & 16.4 & 27.7 & 24.3 \\
        Co-CGE~\cite{mancini2022learning}& & 22.1 & 40.3 & 57.8 & 43.5 & 0.5 & 3.3 & 31.2 & 2.2 & 5.1 & 17.2 & 27.0 & 25.4 \\
        PBadv~\cite{li2024contextual}& &27.7 &44.6 &64.9 &52.8 &1.1 &6.4 &34.2 &4.1 &4.3 &15.3 &37.7 &13.4\\
        ADE~\cite{hao2023learning}& &27.1&44.8&62.4&50.7&1.4&7.6&35.1&4.8&-&-&-&-\\
        HPL~\cite{wang2023hierarchical}& &24.6 &40.2 &63.4 &48.1 &1.37 &7.5 &30.1 &5.8 &6.9 &19.8 &46.4 &18.9\\
        OADis~\cite{Saini_2022_CVPR}& & 25.4 & 41.7 & 58.7 & 53.9 & 0.7 & 4.2 & 33.0 & 2.6 & 5.1 & 16.7 & 26.2 & 24.2 \\
        COT~\cite{kim2023hierarchical}&  &25.0 &41.5 &59.7 &50.3 &1.02 &5.6 &34.4 &4.0 &2.97 &12.1 &36.5 &11.2\\
        \hline
        HDA-OE& ViT &                          \textbf{28.9}&\textbf{45.7}&60.8&\textbf{54.9}&\textbf{2.3}&\textbf{9.8}&\textbf{38.8}&\textbf{7.2}&\textbf{8.1}&\textbf{22.0}&\textbf{40.2}&\textbf{27.6}\\
        \hline
        \end{tabular} }
    \caption{Open-world results on three datasets. Different from close-world setting, open-world setting considers all
    possible compositions in testing.}
\label{tableopen}
\end{table*}

\subsection{Quantitative Result}

\subsubsection{Closed-World CZSL}

The closed-world settings on the test sets of all three datasets are shown in Table \ref{tableclose}. All results are from the respective published papers, and the backbone networks of the compared methods include Resnet and Vit to ensure fair and diverse comparisons. As depicted in Table \ref{tableclose}, our model surpasses other algorithms in terms of AUC, HM, S, and U metrics across all MIT-States and C-GQA datasets. Particularly noteworthy is our model's remarkable performance on the MIT-States dataset. For instance, compared to ADE's 7.4\% AUC, our model achieves 10.6\%, outperforming by more than one-third, which represents a highly substantial advancement. Additionally, on the more challenging C-GQA dataset, we observe a substantial increase in AUC, from 5.2\% to 6.8\%. This highlights our model's robustness to the bias of unseen test compositions.
On the UT-Zappos dataset, our model also attains the best AUC and HM, elevating AUC from 37.7\% to 38.4\% and HM from 52.1\% to 54.0\%. Compared with other models, our model demonstrates superior performance and generalization ability, further substantiating its effectiveness and superiority in closed-world scenarios.

\subsubsection{Open-World CZSL}
Table \ref{tableopen} illustrates the results in the challenging OW-CZSL setting. We observe a significant drop in performance for each method compared to CW-CZSL, particularly on the best unseen class metric, mainly due to the presence of numerous distractors. However, our model consistently outperforms or is on par with all competitors across all metrics. On both the MIT-States and C-GQA datasets, our model achieves notable improvements in AUC, HM, S, and U metrics. Our performance on the MIT-States dataset improves from 5.1\%, 17.2\%, 37.7\%, and 25.4\% to 8.1\%, 22.0\%, 40.2\%, and 27.6\%, respectively. Similarly, on the C-GQA dataset, it increases from 1.4\%, 7.6\%, 35.1\%, and 4.8\% to 2.3\%, 9.8\%, 38.8\%, and 7.2\%, respectively. This underscores our model's robustness to label noise even in the OW-CZSL setting.
Regarding the UT-Zappos dataset, although the performance gap between us and other methods is narrower, we still achieve improvements. This might be attributed to the majority of components in UT-Zappos being feasible. Nevertheless, our model demonstrates enhancements in AUC, HM, and U metrics from 27.7\%, 44.8\%, and 54.7\% to 28.9\%, 45.7\%, and 54.9\%, respectively.
In conclusion, our approach demonstrates exceptional performance and generalization capacity in managing unfamiliar categories and uncontrolled situations, hence confirming its efficacy and superiority in practical contexts.

\subsection{Ablation Studies}

\subsubsection{Impact of the Loss}
\begin{table}[!t]
    \centering
    \resizebox{\linewidth}{!}{
        \begin{tabular}{c|cccccc}
        \hline
        \multirow{2}{*}{Loss}&\multicolumn{6}{c}{UT-Zappos 50K}\\
        \cline{2-7}
        &AUC&	HM	&S&	U&	A&	O\\
        \hline
        $\mathcal{L}_{base}$       &36.6&52.0&62.3&67.9&47.7&74.9 \\
        $\mathcal{L}_{base}$ + $\mathcal{L}_{ea}$            &37.4&51.7&61.5&\textbf{70.8}&\textbf{50.7}&\textbf{77.0}\\
        $\mathcal{L}_{base}$ + $\mathcal{L}_{eo}$              &37.5&52.9&61.6&69.9&50.2&76.7 \\
        $\mathcal{L}_{base}$ + $\mathcal{L}_{ec}$           &37.6&52.3&61.7&70.4&50.4&76.5\\
        $\mathcal{L}_{base}$ + $\mathcal{L}_{ea}$ + $\mathcal{L}_{eo}$         &38.0&53.5&62.0&69.8&49.8&76.7\\
        $\mathcal{L}_{base}$ + $\mathcal{L}_{ea}$ + $\mathcal{L}_{eo}$ + $\mathcal{L}_{ec}$      & \textbf{38.4}&\textbf{54.0}&\textbf{63.4}&68.7& 49.2 & 76.2\\
        \hline
        \end{tabular}
    }
    \caption{We demonstrate quantitatively that our proposed architecture helps disentangle and combine these seen and unseen pairs.}
\label{tableloss}
\end{table}

To study the role of classification attributes, objects, and components modules in our model, we conducted ablation experiments.
%We compared the following model variations: a baseline model with no additional modules ($\mathcal{L}_{cls}$), models with added attributes ($\mathcal{L}_{att}$), objects ($\mathcal{L}_{obj}$), components ($\mathcal{L}_{comp}$), attributes and objects combined ($\mathcal{L}_{att}$ + $\mathcal{L}_{obj}$), and all three combined ($\mathcal{L}_{att}$ + $\mathcal{L}_{obj}$ + $\mathcal{L}_{comp}$), all under the same parameter settings.
These experiments are conducted on the UT-Zappos dataset with the same parameter settings. As shown in Table \ref{tableloss}.
%The results indicate that incorporating attributes, objects, and components improves performance in terms of Harmonic Mean (HM) and Area Under the Curve (AUC).
The baseline model ($\mathcal{L}_{base}$) without any additional modules showed the poorest performance. Adding components ($\mathcal{L}_{ec}$) yielded more significant improvements compared to adding objects ($\mathcal{L}_{eo}$), although the difference in effectiveness was marginal. The combination of all three modules ($\mathcal{L}_{ea}$ + $\mathcal{L}_{eo}$ + $\mathcal{L}_{ec}$) produced the best results, with improvements in AUC and HM of 4.9\% and 3.8\%, respectively.
These findings suggest that learning classification attributes, objects, and components enhances our model's performance by better disentangling and composing seen and unseen pairs. The proposed architecture and the differences in loss functions contribute to this optimization.

\subsubsection{Impact of the temperature parameter $\tau$}
\begin{figure}[!t]
\centering
    \includegraphics[width = 1.0\linewidth]{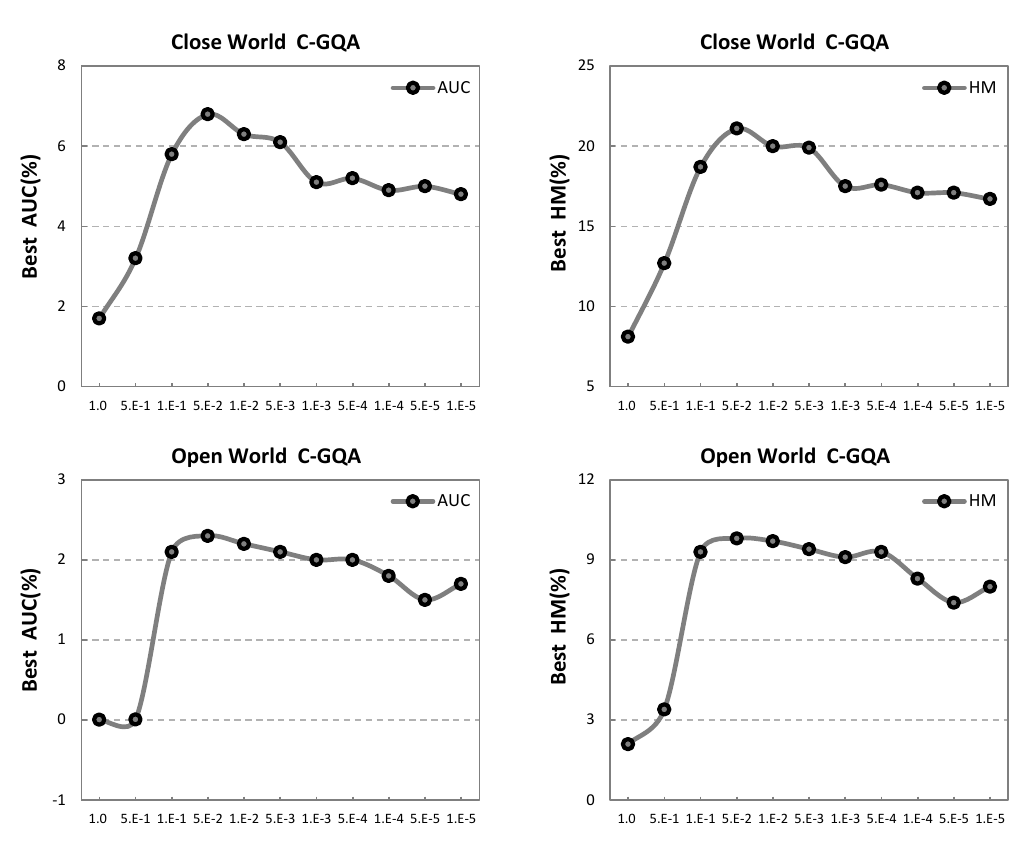}
    \caption{The impact of temperature parameter $\tau$ on the best AUC and HM on the C-GQA dataset in the open and closed world.}
\label{temp}
\end{figure}

The Figure \ref{temp} show the effect of varying the temperature parameter on the AUC and HM  performance metrics in both Close World and Open World settings on the C-GQA dataset. As observed, both metrics initially improve as the temperature parameter decreases from 1.0, reaching optimal values when the temperature is around 0.05. Specifically, for the CW setting, the AUC peaks at around 6\%, and the HM reaches close to 20\%, while in the OW setting, the AUC peaks around 2\%, and the HM approaches 10\%. After this optimal point (at approximately  $\tau = 0.05$ ), further reductions in the temperature lead to a decline in both AUC and HM values.
The chosen temperature of $\tau = 0.05$ thus strikes a balance, maximizing the model's performance across both settings. This parameter seems to enhance the model's ability to differentiate and generalize across attributes effectively, likely due to the fine-tuning of similarity calculations in the embedding space.

\subsubsection{Impact of the dataset hybrid strategy}
\begin{table}[!t]
    \centering
    \resizebox{0.9\linewidth}{!}{
        \begin{tabular}{c|ccc|cccc}
        \hline
        \multirow{2}{*}{Models}&\multicolumn{3}{c|}{Strategie}&\multicolumn{4}{c}{UT-Zappos 50K}\\
        \cline{2-8}
        &att&obj&maa&AUC&	HM	&S&	U\\
        \hline
        $M_{1}$          & & & &29.6&45.3&57.0&63.2\\
        $M_{2}$      &\checkmark & & &28.3&44.7&54.4&61.2\\
        $M_{3}$      & &\checkmark& & \textbf{38.4}&\textbf{54.0}&\textbf{63.4}&\textbf{68.7}\\
        $M_{4}$  &\checkmark&\checkmark& &29.3&45.0&56.8&60.1\\
        $M_{5}$  & &\checkmark&\checkmark&27.4&44.5&55.5&58.0\\
        \hline
        \end{tabular}
    }
    \caption{Ablation study of different datasets expansion on UT-zappos dataset.
Base represents no data enhancement.}
\label{tabledataset}
\end{table}
\begin{figure*}[!t]
    \includegraphics[width = \textwidth]{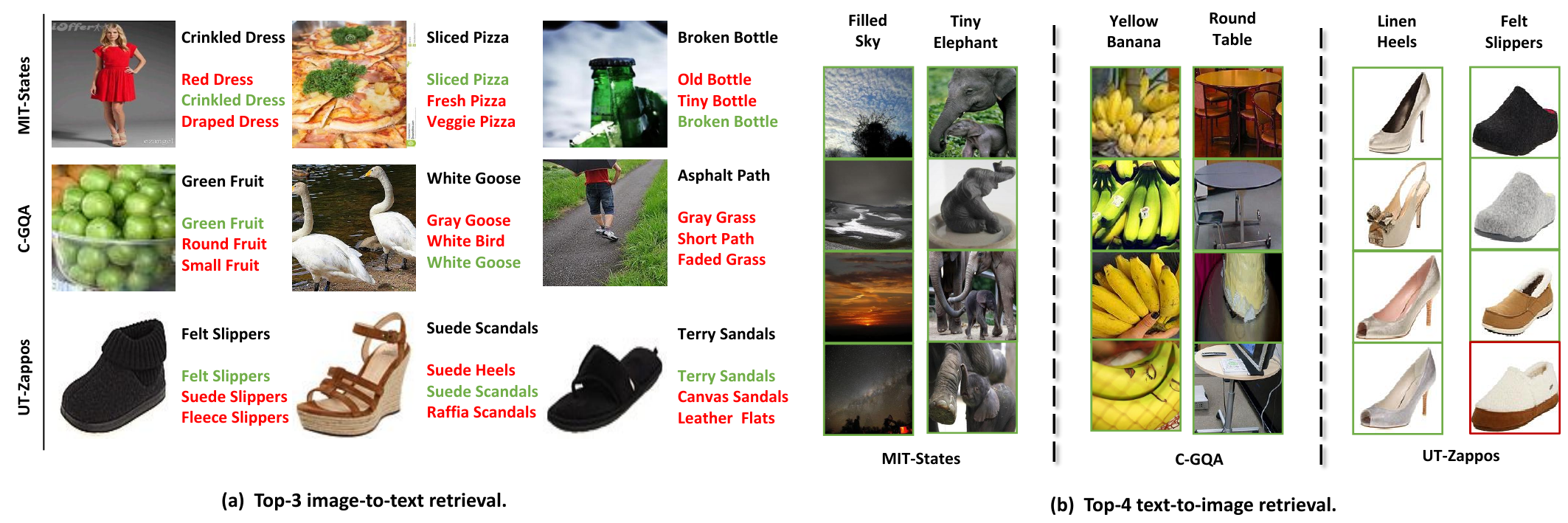}
    \centering
    \caption{Qualitative Result. (a) Each image has a ground truth label (black text) and 5 retrieval results (colored text), where the green text is the correct prediction. (b) In the last row ``Felt Slipper", the wrong image (red box) is ``fleece Slippers".}
\label{figureimage}
\end{figure*}
%For database augmentation, we use images of the same object but with different attributes to connect with the original image in the model, creating a new database.
In order to create a new database for database augmentation, we connect photographs of the same object in the model with various attributes. This approach is called the \textit{obj} joining strategy. To verify its effectiveness, we compared several strategies in Table \ref{tabledataset}: model $M_{1}$ without data augmentation, model $M_{2}$ using the \textit{att} joining strategy (connecting images of the same attributes but different objects with the original image), model $M_{3}$ using the \textit{obj} joining strategy, model $M_{4}$ using the \textit{att+obj} joining strategy, and model $M_{5}$ using the \textit{MAA }strategy \cite{kim2023hierarchical}, which combines the \textit{obj} joining strategy with a data augmentation method.
%The MAA strategy generates virtual attribute features by mixing a few attributes to balance the long-tail distribution of the data.
As seen in Table \ref{tabledataset}, model $M_{2}$, which uses the \textit{att} joining strategy, performs worse than model $M_{1}$, reducing the model's accuracy for both visible and invisible components.
%(the accuracy of visible components drops from 57.0 to 54.4, and for invisible components, it drops from 63.2 to 61.2).
Similarly, model $M_{4}$ does not perform as well as model $M_{3}$, likely due to confirmation bias. In contrast, model $M_{3}$, which uses the \textit{obj} joining strategy, shows significant improvements: AUC increases by 8.8\%, HM by 8.7\%, S by 6.4\%, and U by 5.5\%, outperforming the base model $M_{1}$ and excelling in identifying both visible and invisible pairs. %Overall, model $M_{3}$ outperformed all other models in every indicator.
For models $M_{3}$ and $M_{5}$, we found that model $M_{5}$, which incorporates the \textit{MAA} strategy, performed significantly worse than model $M_{3}$ and even lower than the base model $M_{1}$. This suggests that the \textit{MAA} strategy may have issues with underfitting during data mixing.
%In summary, the \textit{obj} joining strategy effectively improves model performance, whereas the \textit{att} joining strategy and \textit{MAA} strategy did not provide the expected improvements and, in some cases, even reduced the model's accuracy.
In conclusion, the \textit{att} joining approach and \textit{MAA} method did not produce the anticipated advantages and even decreased the model's accuracy. In contrast, the \textit{obj} joining strategy effectively enhances model performance.

\subsection{Image Retrieval}
In this section, we present qualitative results for new compositions using image-to-text retrieval. Given an image, we retrieve the three closest text composition embeddings. The top three predictions on the UT-Zappos, MIT-States, and C-GQA datasets are shown in Figure \ref{figureimage} (a). Our model correctly predicts the top three results in most cases.
For the image labeled as ``Crinkled Dress" in MIT-States, our model first predicts its attribute as red, as it is difficult to focus on a specific attribute of the dress due to its multiple attributes.
%Additionally, the images in the training set are restricted in terms of object states. For example,
The training images of ``Path" in the C-GQA dataset hardly present the attribute of ``Asphalt", causing our model to incorrectly classify the ``Path" labeled as ``Asphalt Path" in the image as ``Grass". Therefore, for the identified object ``Grass", the model can only focus on the attributes conditioned on ``Grass" and find appropriate attributes to match the image. We point out that this failure is partly attributed to the incomplete annotation problem. The multi-label nature of natural images provides additional challenges for the CZSL task.
Then, we consider text-to-image retrieval. In Figure \ref{figureimage} (b), we retrieve the top four closest visual features based on feature distance on the UT-Zappos, MIT-States, and C-GQA datasets. We can observe that in most cases, the retrieved images are correct. One exception is when retrieving ``Felt Slipper", where the third closest image is ``Fleece Slippers". Although ``Felt Slipper" and ``Fleece Slippers" are not the same composition, they are quite similar visually.
The image and text retrieval experiments verify that our model effectively embeds visual features and words into a unified space.

\section{Conclusion}
In this paper, we propose a Hybrid Discriminative Attribute-Object Embedding (HDA-OE) network to solve CZSL task. 
We hypothesize that complex interdependencies between subclasses in attribute-object combinations influence visual feature differences. By introducing a subclass-focused embedding expert module, we reveal and leverage these fine-grained interdependencies, enhancing the model's ability to generalize to unseen categories. 
To address critical challenges such as the high degree of hybridity and the long-tail distribution of real-world image features, we introduce an attribute-driven data synthesis. This strategy integrates feature information from multiple databases, thereby improving the model's recognition accuracy and robustness when handling diverse and rare combinations.
We validate the effectiveness of our method on three challenging datasets. Comparative experimental results demonstrate that our approach outperforms previous state-of-the-art methods.

%We hypothesize that complex interdependence structures exist between subclasses. By leveraging an embedding expert module focused on subclass differentiation, we can uncover and utilize these interdependencies, generalizing them to unseen classes. This significantly enhances the model's capacity to learn and distinguish between different concept combinations.
%To address key challenges such as the high degree of hybridity and the long-tail distribution of real-world image features, we introduce a new dataset mixing strategy. This strategy improves the recognition accuracy and robustness of the model by integrating feature information from multiple databases.
{
    \small
    \bibliographystyle{ieeenat_fullname}
    \bibliography{main}
}

\end{CJK}
\end{document}